\documentclass[11pt,a4paper]{article}
\usepackage{times,latexsym}
\usepackage{url}
\usepackage[T1]{fontenc}
\usepackage{graphicx}
\usepackage{amsmath}
\usepackage[table,xcdraw]{xcolor}
\usepackage{dsfont}
\usepackage{booktabs,arydshln}
\usepackage{microtype}
\usepackage{amssymb}
\newtheorem{theorem}{Theorem}

\usepackage{footnote}
\makesavenoteenv{tabular}
\makesavenoteenv{table}

\usepackage{enumitem}
\setlist{leftmargin=*}

\usepackage[acceptedWithA]{tacl2021v1}

\usepackage{xspace,mfirstuc,tabulary}

\newif\iftaclinstructions
\taclinstructionsfalse 
\iftaclinstructions
\renewcommand{\confidential}{}
\renewcommand{\anonsubtext}{(No author info supplied here, for consistency with
TACL-submission anonymization requirements)}
\newcommand{\instr}
\fi

\iftaclpubformat 

\else

\fi

\title{Conformalizing Machine Translation Evaluation}

\author{
Chrysoula Zerva$^{1, 3}$ \quad André F. T. Martins$^{1, 2, 3}$ 
\\
$^1$Instituto de Telecomunicações \quad\quad\quad $^2$Unbabel \quad  \\
$^3$Instituto Superior Técnico \& LUMLIS (Lisbon ELLIS Unit) \\
{\small \texttt{\{chrysoula.zerva, andre.t.martins\}@tecnico.ulisboa.pt}}\\
}

\date{}

\begin{document}
\maketitle
\begin{abstract}
Several uncertainty estimation methods have been recently proposed for machine translation evaluation. While these methods can provide a useful indication of when not to trust model predictions, we show in this paper that the majority of them tend to  underestimate model uncertainty, and as a result they often produce misleading confidence intervals that do not cover the ground truth.  
We propose as an alternative the use of \textit{conformal prediction}, a distribution-free method to obtain confidence intervals with a theoretically established guarantee on coverage. First, we demonstrate that split conformal prediction can ``correct'' the confidence intervals of previous methods to yield a desired coverage level. Then, we highlight biases in estimated confidence intervals, both in terms of the translation language pairs and the quality of translations. We apply conditional conformal prediction techniques to obtain calibration subsets for each data subgroup, leading to \textit{equalized} coverage.
\end{abstract}

\section{Introduction}
Neural models for natural language processing (NLP) are able to tackle increasingly challenging tasks with impressive performance. 
However, their deployment in real world applications does not come without risks. 
For example, systems that generate fluent text might mislead users with fabricated facts, particularly if they do not expose their confidence. 
High performance does not guarantee an accurate prediction for every instance---and even less so when an instance is noisy or out of distribution---which makes uncertainty quantification methods more important than ever. 

\begin{figure}[t!]
    \centering
\begin{minipage}{\columnwidth}
    \includegraphics[width=\columnwidth]{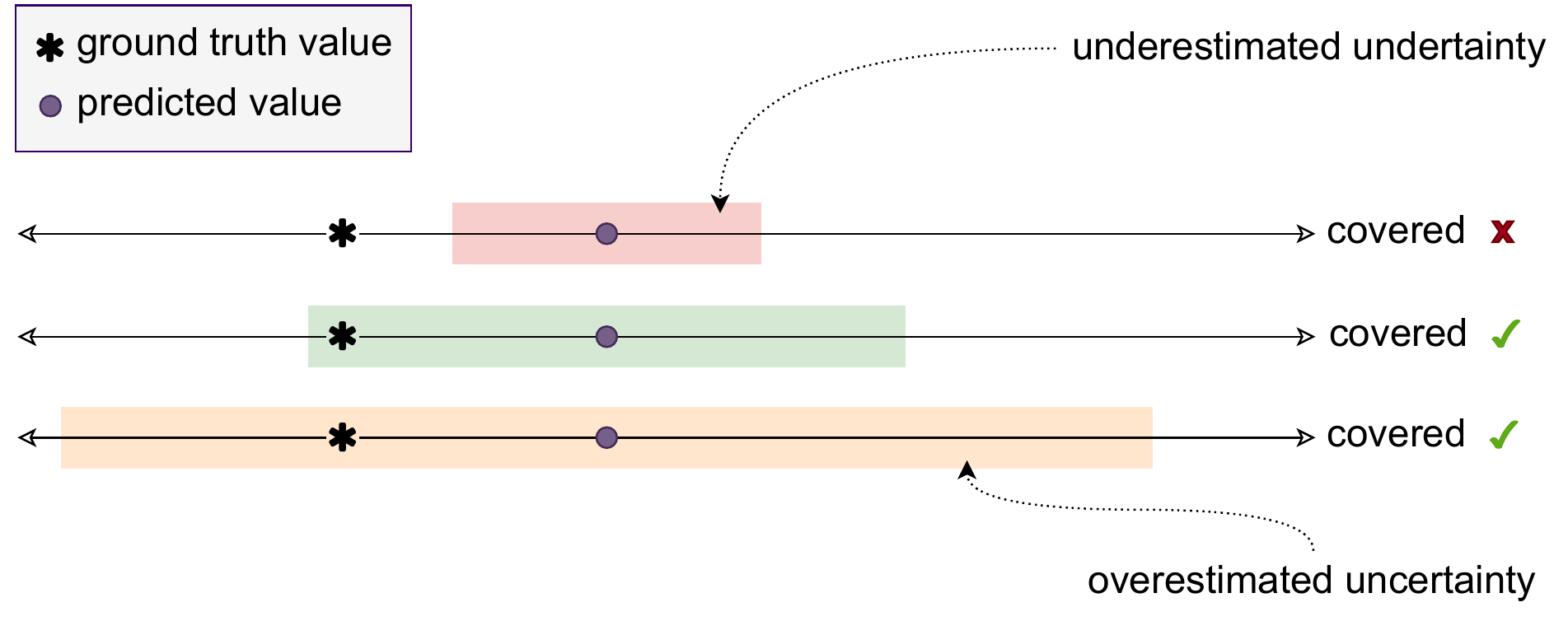}
    \caption{Predicted confidence intervals and coverage for the same ground truth/prediction points. We consider the middle (green) interval to be desired as it covers the ground truth but does not overestimate the model uncertainty.}
    \label{fig:cover_ideal}
\end{minipage}\\\vspace{0.5cm}%
\begin{minipage}{\columnwidth}
    \includegraphics[width=\columnwidth]{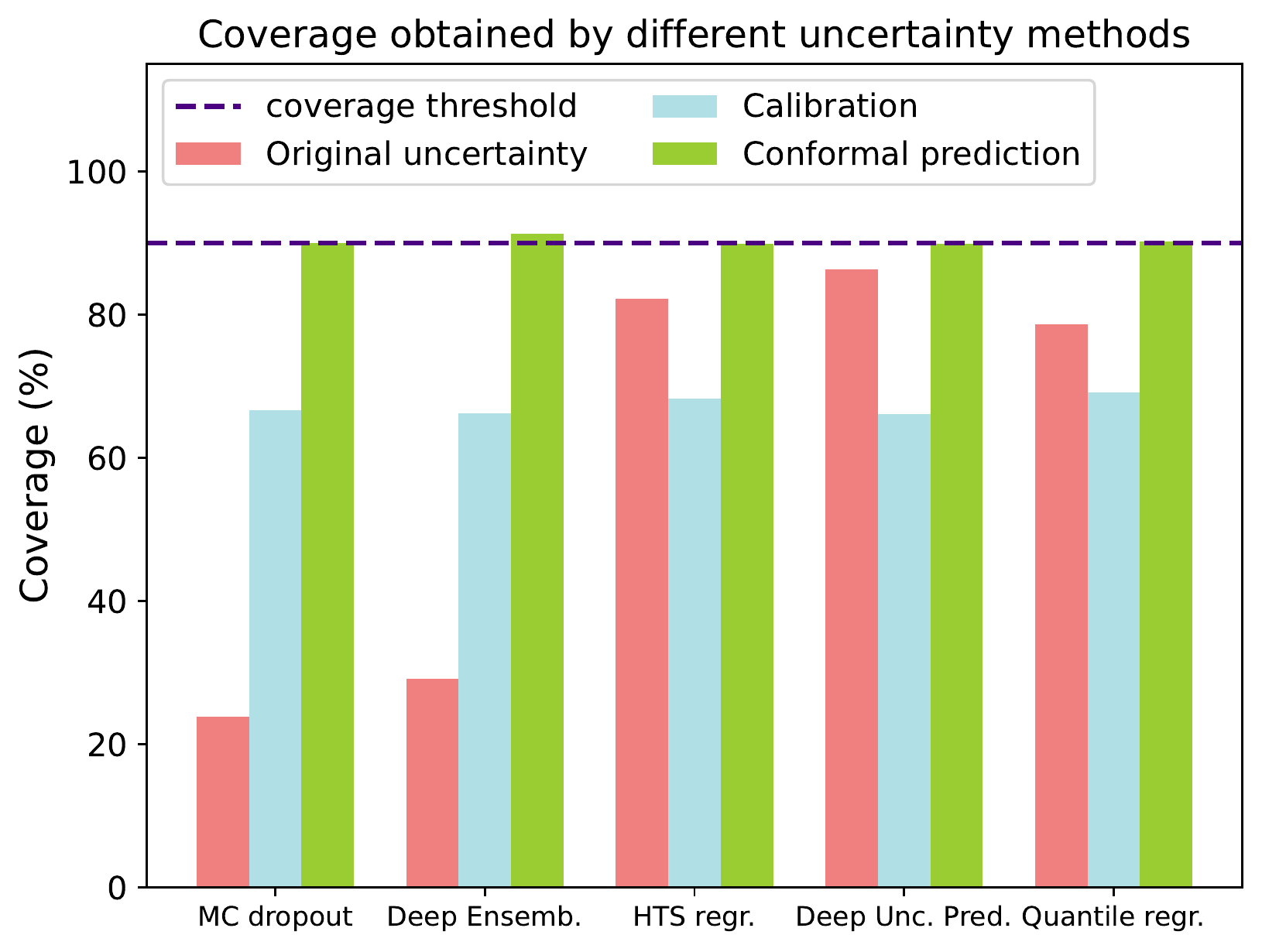}
    \caption{Coverage obtained by different uncertainty predictors. We compare originally obtained values (red), with values after calibration (light blue) and after conformal prediction (green) with the desired coverage threshold (dashed line) set to 0.9 (90\%).}
    \label{fig:cover_ex}
\end{minipage}
    \end{figure}

While most work on uncertainty estimation for NLP has focused on classification tasks, uncertainty quantification for text regression tasks, uncertainty quantification has recently gained traction for text regression tasks, such as  machine translation (MT) evaluation, semantic sentence similarity, or sentiment analysis \cite{wang2022uncertainty,glushkova2021uncertainty}.  
A wide range of methods have been proposed for estimating uncertainty in regression, the majority of which involve underlying assumptions on the distribution or the source of uncertainty \cite{kendall2017uncertainties,kuleshov2018accurate,amini2020deep,ulmer2023prior}. However, it has been shown that such assumptions are often unrealistic and may lead to misleading results \cite{izmailov2021dangers,zerva2022disentangling}. 
More importantly, \textbf{most commonly used methods provide confidence intervals without any theoretically established guarantees with respect to coverage}. 
In other words, while a representative confidence interval should include (cover) the ground truth target value for each instance (and ideally the bound of the confidence interval should be close to the ground truth as shown in Figure~\ref{fig:cover_ideal}), frequently the predicted interval is much narrower, excluding the ground truth and underestimating the model uncertainty.
In fact, for the concrete problem of MT evaluation,  we show that \textbf{the majority of uncertainty quantification methods achieve very low coverage} even after calibration, as can be observed in Figure \ref{fig:cover_ex}. Finally, it has been shown that while uncertainty quantification can shed light on model weaknesses and biases, \textbf{the uncertainty prediction methods themselves can suffer from biases} and provide unfair and misleading predictions for specific data subgroups or for examples with varying levels of difficulty \cite{cherian2023statistical,Ding_2020_CVPR_Workshops,bostrom2020mondrian}.

To address the aforementioned shortcomings,  we explore the use of \textbf{conformal prediction} as a means to obtain more trustworthy confidence intervals on textual regression tasks, using MT evaluation as the primary paradigm. We rely on the fact that given a scoring or uncertainty estimation function, conformal prediction can provide statistically rigorous uncertainty intervals for regression models \cite{angelopoulos2021gentle,vovk2005algorithmic,vovk2022conformal}. More importantly, the conformal prediction methodology provides theoretical guarantees about coverage over a test set, given a chosen coverage threshold. The predicted uncertainty intervals are thus valid in a \textbf{distribution-free} sense:
they possess explicit, non-asymptotic guarantees even without distributional assumptions or model assumptions \cite{angelopoulos2021gentle,vovk2005algorithmic}. 


We specifically show that previously proposed uncertainty quantification methods can be used to design non-conformity scores for split conformal prediction \cite{papadopoulos2008inductive}. We demonstrate that, regardless of the initially obtained coverage, the application of conformal prediction can increase coverage to the desired ---user defined--- value (see Figure \ref{fig:cover_ex}). To this end, we compare four parametric uncertainty estimation methods (Monte Carlo dropout, deep ensembles, heteroscedastic regression,  and direct uncertainty prediction) and one non-parametric method (quantile regression) with respect to coverage and distribution of uncertainty intervals. 

Moreover, we investigate the fairness of obtained intervals for two different attributes: (1) translation language pair; and (2) translation difficulty, as reflected by human quality estimates. We highlight unbalanced coverage for both cases and demonstrate how \textbf{conditional conformal prediction} \cite{angelopoulos2021gentle,bostrom2020mondrian,bostrom2021mondrian} can address such imbalances effectively.

\section{Conformal Prediction}
\label{sec:conformal}
In this section, we provide background on conformal prediction and introduce the notation used throughout this paper. Later in \S\ref{sec:conformal_mt_eval} we show how this framework can be used for uncertainty quantification in MT evaluation.

\subsection{Desiderata}

Let $X \in \mathcal{X}$ and $Y \in \mathcal{Y}$ be random variables representing inputs and outputs, respectively; in this paper we focus on regression, where $\mathcal{Y} = \mathbb{R}$. We use upper case to denote random variables and lower case to denote their specific values. 
Traditional machine learning systems use training data to learn predictors $\hat{y}: \mathcal{X} \rightarrow \mathcal{Y}$ which, when given a new test input $x_\mathrm{test}$, output a point estimate $\hat{y}(x_\mathrm{test})$. However, such point estimates lack  uncertainty information. 
Conformal prediction \cite{vovk2005algorithmic}, in contrast, considers \textit{set} functions  
$\mathcal{C} : \mathcal{X} \rightarrow 2^\mathcal{Y}$, 
providing the methodology for, given $x_\mathrm{test}$, returning a \textbf{prediction set}  $\mathcal{C}(x_\mathrm{test}) \subseteq \mathcal{Y}$ 
with theoretically established guarantees regarding the coverage of the ground truth value. For regression tasks, this prediction set is usually a confidence interval (see Figure \ref{fig:cover_ideal}). 
Conformal prediction techniques have recently proved  useful in many applications: for example, in the U.S. presidential election in 2020, the Washington Post used conformal prediction to estimate the number of outstanding  votes \citep{cherian2020washington}. 

Given a desired confidence level (e.g. 90\%), these methods have a formal  guarantee that, in expectation, $\mathcal{C}(X_\mathrm{test})$ contains the true value $Y_\mathrm{test}$ with a probability equal to or higher than the given confidence level. 
Importantly, this is done in a \textit{distribution-free} manner, i.e., without making any assumptions on the data distribution beyond \textbf{exchangeability}, a weaker assumption than  independent and identically distributed (i.i.d.) data.%
\footnote{\label{footnote:exchangeable}Namely, the data distribution is said to be \textit{exchangeable} iff, for any sample $(X_i, Y_i)_{i=1}^n$ and any permutation function $\pi$, we have $\mathds{P}((X_{\pi(1)}, Y_{\pi(1)}), \ldots, (X_{\pi(n)}, Y_{\pi(n)})) = \mathds{P}((X_1, Y_1), \ldots, (X_n, Y_n))$. 
If the data distribution is i.i.d., then it is automatically exchangeable, since $\mathds{P}((X_1, Y_1), \ldots, (X_n, Y_n)) = \prod_{i=1}^n \mathds{P}(X_i, Y_i)$ and the product of scalars is commutative.  
By de Finetti's theorem \citep{de1929funzione}, exchangeable observations are conditionally independent relative to some latent variable.} %

In this paper, we use a simple  inductive method called \textit{split} conformal prediction  \cite{papadopoulos2008inductive}, which requires the following ingredients: 
\begin{itemize}
    \item A mechanism to obtain \textbf{non-conformity scores} $s(x,y)$ for each instance, i.e., a way to estimate how ``unexpected'' an instance is with respect to the rest of the data. 
    In this work, we do this by leveraging a pretrained predictor $\hat{y}(x)$ together with \textit{some} heuristic notion of uncertainty---our method is completely agnostic about which model is used for this. We describe in \S\ref{sec:nonconf_scores} the non-conformity scores we design in our work.
    \item A held-out \textbf{calibration set} containing $n$ examples, $\mathcal{S}^\mathrm{cal} = \{(x_1,y_1), \ldots,(x_n,y_n)\}$. The underlying distribution from which the calibration set is generated is assumed unknown but it must be exchangeable (see footnote~\ref{footnote:exchangeable}). 
    \item A desired \textbf{error rate} $\alpha$ (e.g. $\alpha=0.1$), such that the coverage level will be $1-\alpha$ (e.g. 90\%). 
\end{itemize}
These ingredients are used to generate prediction sets 
for new test inputs. 
Specifically, let $(s_1, \ldots, s_n)$ be the non-conformity scores of each example in the calibration set, i.e., $s_i := s(x_i, y_i)$. 
Define $\hat{q}$ as the ${\lceil(n+1)(1-\alpha)\rceil}/{n}$ \textbf{empirical quantile} of these non-conformity scores, where $\lceil \cdot \rceil$ is the ceiling function. This quantile can be easily obtained by sorting the $n$ non-conformity scores and examining the tail of the sequence. %
Then, for a new test input $x_\mathrm{test}$, we 
output the prediction set 
\begin{equation}\label{eq:prediction_set}
C_{\hat{q}}(x_\mathrm{test}) = \{y \in \mathcal{Y}\,:\, s(x_\mathrm{test}, y) \le \hat{q} \}. 
\end{equation}
We say that \textbf{coverage} holds if the true output $y_\mathrm{test}$ lies in the prediction set, i.e.,  
$y_\mathrm{test} \in C_{\hat{q}}(x_\mathrm{test})$. 
This simple procedure has the following theoretical coverage guarantee: 
\begin{theorem}[\citealt{vovk1999machine,vovk2005algorithmic}] \label{theo:conf}
Using the above quantities, the following bounds hold: 
\begin{equation*}
    \mathds{P} \big(Y_\mathrm{test} \in \mathcal{C}_{\hat{q}}(X_\mathrm{test})\big) \in \left[1-\alpha, \,\, 1-\alpha + \frac{1}{n+1}\right].
    \label{eq:conf}
\end{equation*}
\end{theorem} 
This result tells us two important things: (i) the expected coverage is \textit{at least} $1-\alpha$, and (ii) with a large enough calibration set (large $n$), the procedure outlined above does not overestimate the coverage too much, so we can expect it to be \textit{nearly} $1-\alpha$.%
\footnote{For most purposes, a reasonable size for the calibration set is $n\approx 1000$. See \citet[\S 3.2]{angelopoulos2021gentle}.} %



\subsection{Non-conformity scores}\label{sec:nonconf_scores}

Naturally, the result stated in  Theorem~\ref{theo:conf} is only practically useful if the prediction sets $\mathcal{C}_{\hat{q}}(X_\mathrm{test})$ are small enough to be informative---to ensure this, we need a good heuristic to generate the non-conformity scores. 
In this paper, we are concerned with regression problems ($\mathcal{Y} = \mathbb{R}$), so we define the prediction sets to be \textbf{confidence intervals}. 
We assume we have a pretrained regressor $\hat{y}(x)$, and we consider two scenarios, one where we generate \textit{symmetric} intervals (i.e., where $\hat{y}(x)$ is the midpoint of the interval) and a more general scenario where intervals can be \textit{non-symmetric}. 

\paragraph{Symmetric intervals.} 
In this simpler scenario, we assume that, along with $\hat{y}(x)$, we have a corresponding uncertainty heuristic $\delta(x)$, where higher $\delta(x)$ values signify higher uncertainty. 
An example---to be elaborated upon in \S\ref{sec:parametric}---is where $\delta(x)$ is the quantile of a symmetric probability density, such as a Gaussian, which can be computed analytically from the variance.  
We then define the non-conformity scores as  
\begin{equation}
    s(x,y) = \frac{| y - \hat{y}(x)|}{\delta(x)}
        \label{eq:score}
\end{equation}
and follow the procedure above to obtain the quantile $\hat{q}$ from the calibration set. 
Then, for a random test point $(X_\mathrm{test},Y_\mathrm{test})$ and from \eqref{eq:conf} and \eqref{eq:score}, we have:
\begin{equation}\label{eq:conf2}
\mathds{P} \big[|Y_\mathrm{test}-\hat{y}(X_\mathrm{test})| \leq {\delta(X_\mathrm{test})}  \hat{q}\big] \gtrsim 1-\alpha,
\end{equation}
which corresponds to the confidence interval 
\begin{equation}
    \mathcal{C}_{\hat{q}}(x) = \big[\hat{y}(x) - \hat{q}\delta(x), \,\, \hat{y}(x) + \hat{q}\delta(x)\big].
    \label{eq:confint}
\end{equation}
We examine this procedure in \S\ref{sec:parametric} with various uncertainty heuristics (Monte Carlo dropout, deep ensembles, heteroscedastic regression, and direct uncertainty prediction estimates). 

\paragraph{Non-symmetric intervals.} 
Sometimes, better heuristics can be obtained which are non symmetric, i.e., where there is larger uncertainty in one of the sides of the interval---we will see a concrete example in \S\ref{sec:quantile} where we describe a non-parametric quantile regression procedure (although this might happen as well with parametric heuristics based on fitting non-symmetric distributions, such as the skewed beta distribution). 
In this case, we assume left and right uncertainty estimates $\delta_-$ and $\delta_+$, both non-negative and satisfying $\delta_- \le \delta_+$, 
and define the non-conformity scores as:
\begin{align}\label{eq:scores_nonsym}
    s(x,y) &= \left\{
    \begin{array}{ll}
    \frac{y-\hat{y}(x)}{\delta_+(x)} & \text{if $y \ge \hat{y}(x)$}\\
    \frac{\hat{y}(x)-y}{\delta_-(x)} & \text{if $y < \hat{y}(x)$}.
    \end{array}
    \right.
\end{align}
This leads to prediction sets 
\begin{align}\label{eq:confint_nonsym}
    C_{\hat{q}}(x) &= \big[\hat{y}(x) - \hat{q}\delta_-(x), \,\,\hat{y}(x) + \hat{q}\delta_+(x)\big],
\end{align}
which also satisfy Theorem~\ref{theo:conf}. 
Naturally, when $\delta_- = \delta_+ := \delta$, this procedure recovers the symmetric case.


\section{Conformal MT Evaluation}\label{sec:conformal_mt_eval}

\begin{figure*}[!htb]
    \centering
    \includegraphics[width=\linewidth]{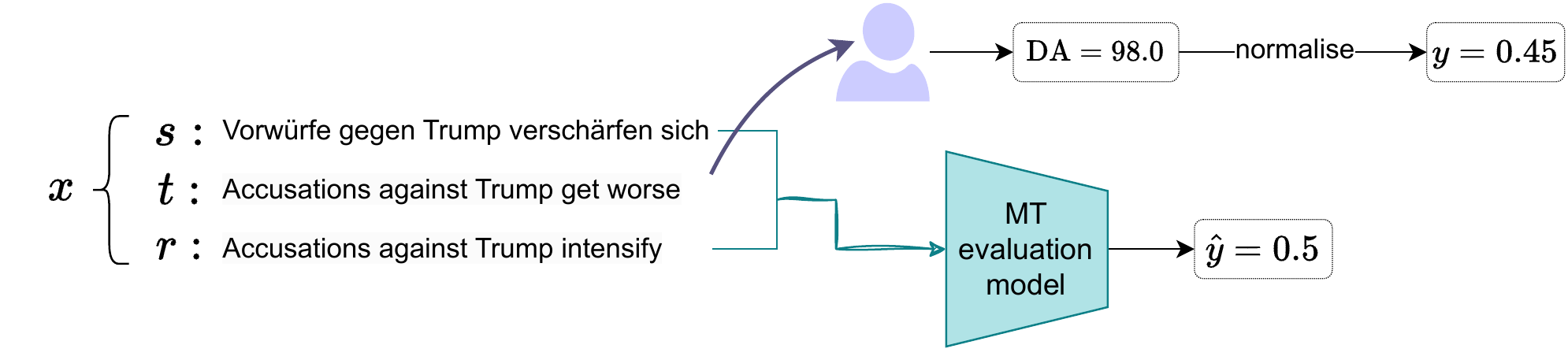}
    \caption{Example of MT evaluation instance.}
    \label{fig:mt_example}
\end{figure*}
We now apply the machinery of conformal prediction to the problem of MT evaluation, experimenting with a range of uncertainty quantification heuristics to generate $\hat{y}(x)$ and $\delta(x)$ (or $\delta_-(x)$ and $\delta_+(x)$ in the non-symmetric case). 

In MT evaluation, the input is a triplet of source, automatic translation, and human reference segment, $x := \langle s, t, r \rangle$, and the goal is to predict a scalar value $\hat{y}(x)$ that corresponds to the estimated quality of the translation $t$. 
The ground truth is a quality score $y$ manually produced by a human annotator, called a \textit{direct assessment} (DA) \citep{graham2013continuous}. We use DA scores that are standardised for each annotator. An example instance is shown in Figure \ref{fig:mt_example}.

We describe our datasets and experimental setup in \S\ref{sec:data}. 
With the application of symmetric parametric uncertainty methods, described in \S\ref{sec:parametric}, we obtain heuristics $\hat{y}$ and $\delta$ which we use to obtain non-conformity scores via \eqref{eq:score}, leading to the confidence intervals $\mathcal{C}_{\hat{q}}(x)$ in \eqref{eq:confint}, for each $x$ triplet. 
Alternatively, in \S\ref{sec:quantile} we describe a non-symmetric and non-parametric method 
which returns $\hat{y}$, $\delta_-$, and $\delta_+$, and which we will use to compute the non-conformity scores \eqref{eq:scores_nonsym} and confidence intervals \eqref{eq:confint_nonsym}. 




\subsection{Experimental setup}
\label{sec:data}

\paragraph{Model.} We use \textsc{Comet} as the underlying MT quality evaluation model \citep{rei2020comet}. We train the \textsc{Comet} model using a pretrained XLM-RoBERTa-Large encoder fine-tuned for 2 epochs with the default training configurations.%
\footnote{More precisely,  we used the \textit{wmt-large-da-estimator-1719}  available at: \url{https://unbabel.github.io/COMET/html/models.html}.} 

\paragraph{Data.} For training, we use the direct assessment (DA) data from the WMT17-19 metrics shared tasks \cite{ma2018results,ma2019results}. We evaluate our models on the WMT20 metrics dataset \cite{mathur2020results}. For the calibration set $\mathcal{S}^{\mathrm{cal}}$, we use repeated random sub-sampling for $k = 10$ runs. The WMT20 test data includes 16 language pairs, of which 9 pairs are into-English and 7 pairs are out-of-English translations. For the calibration set sub-sampling, we sample uniformly from each language pair. For metrics for which we report averaged performance we use micro-average over all language pairs.


\subsection{Uncertainty quantification methods}
\label{sec:unc_meth} 
We experiment with a diverse set of uncertainty prediction methods, accounting both for parametric and non-parametric uncertainty prediction. We extensively compare all the parametric methods previously used in MT evaluation \citep{zerva2022disentangling}, which return symmetric confidence intervals, and also experiment with \textbf{quantile regression} \cite{koenker2001quantile}, a simple non-parametric approach that has never been used for MT evaluation, and which can return non-symmetric intervals. 

\subsubsection{Parametric uncertainty}\label{sec:parametric}
We compare a set of different parametric methods which fit the quality scores in the training data to an input-dependent Gaussian distribution $\mathcal{N}(\hat{\mu}(x),\,\hat{\sigma}^{2}(x)) $. 
All these methods lead to  \textbf{symmetric} confidence intervals (see Eq.~\ref{eq:confint}). 
We use these methods to obtain estimates $\hat{y}(x) := \hat{\mu}(x)$. Then we use $\hat{sigma}$ to extract the corresponding uncertainty estimates as $\delta(x) := \mathrm{probit}(1-\frac{\alpha}{2}) \hat{\sigma}$, which correspond to the $\frac{\alpha}{2}$ and $1-\frac{\alpha}{2}$ quantiles of the Gaussian, for a given confidence threshold $1-\alpha$. For $\alpha = 0.1$ (i.e., a 90\% confidence level) this results in $\delta(x) = 1.64\times\hat{\sigma}$. We describe the concrete methods used to estimate $\hat{\mu}(x)$ and $\hat{\sigma}(x)$ below.

\paragraph{MC dropout (MCD).} This is a variational inference
technique approximating a Bayesian network with
a Bernoulli prior distribution over its weights \cite{gal2016dropout}. By
retaining dropout layers during multiple inference
runs, we can sample from the posterior distribution over the weights. As such, we can approximate the uncertainty over a test instance $x$ through a Gaussian distribution with the empirical mean $\hat{\mu}(x)$ and variance $\hat{\sigma}^2(x)$ of the quality estimates $\{\hat{y}_1, \ldots, \hat{y}_N\}$. We use 100 runs, following the analysis of \citet{glushkova2021uncertainty}. 

\paragraph{Deep ensembles (DE).} 
This method 
\cite{lakshminarayanan2017simple} trains an ensemble of
neural models with the same architecture but different initializations. During inference, we collect the predictions of each single model and return  $\hat{\mu}(x)$ and $\hat{\sigma}^2(x)$ as in MC dropout. 
We use $N=5$ checkpoints obtained with different initialization seeds, following \citet{glushkova2021uncertainty}.

\paragraph{Heteroscedastic regression (HTS).}
We follow \citet{le2005heteroscedastic} and \citet{kendall2017} and incorporate  $\hat{\sigma}^2(x)$ as part of the training objective. This way, a regressor is trained to output two values: (1) a mean score $\hat{\mu}(x)$ and (2) a variance score $\hat{\sigma}^2(x)$. 
This predicted mean and variance parameterize a Gaussian distribution $\mathcal{N}(y; \hat{\mu}(x; \theta), \hat{\sigma}^2(x; \theta))$, where $\theta$ are the model parameters. 
The negative log-likelihood loss function is used:
\begin{align}
\label{eq:hts1}
\mathcal{L}_{\mathrm{HTS}}(\hat{\mu}, \hat{\sigma}^2; y) &= \frac{(y - \hat{\mu})^2}{2\hat{\sigma}^2} + \frac{1}{2}\log \hat{\sigma}^2. 
\end{align}
This framework is particularly suitable to express aleatoric uncertainty due to heteroscedastic noise, as the framework allows larger variance to be assigned to ``noisy'' examples which will have the effect of downweighting the squared term in the loss. 

\paragraph{Direct uncertainty prediction (DUP).} 
This is a two-step procedure which relies on the assumption that the total uncertainty over a test instance is equivalent to the {generalization error} of the regression model \cite{jain2021deup}. 
A standard regression model $\hat{y}(x)$ is first fit on the training set and then applied to a held-out validation set $\mathcal{S}^{\mathrm{val}}$. Then, a second model is trained on this held-out set to regress on the error $\epsilon = |\hat{y}(x) - y|$ incurred by the first model predictions,  approximating its uncertainty. To train the error predicting model, we follow the setup of \cite{zerva2022disentangling},  using as inputs the $x^{\mathrm{val}} = \langle s, t, \mathcal{R} \rangle$ triplets combined with the predictions ${\hat{y}^{\mathrm{val}}}$ of first model, which are used as bottleneck features in an intermediate fusion fashion. The loss function is  
\begin{align}
   \mathcal{L}_\mathrm{DUP}(\hat{\epsilon}; \epsilon) &= \frac{\epsilon^2}{2 \hat{\epsilon}^2} + \frac{1}{2}\log(\hat{\epsilon})^2.
   \label{eq:loss3}
\end{align} 
We use $\hat{\sigma}(x) = \hat{\epsilon}(x)$ as the uncertainty heuristic. 

\subsubsection{Non-parametric uncertainty: Quantile regression (QNT)}\label{sec:quantile}

\begin{figure}[t]
    \centering
    \includegraphics[trim={5cm 20cm 5cm 3cm},clip,width=0.75\columnwidth]{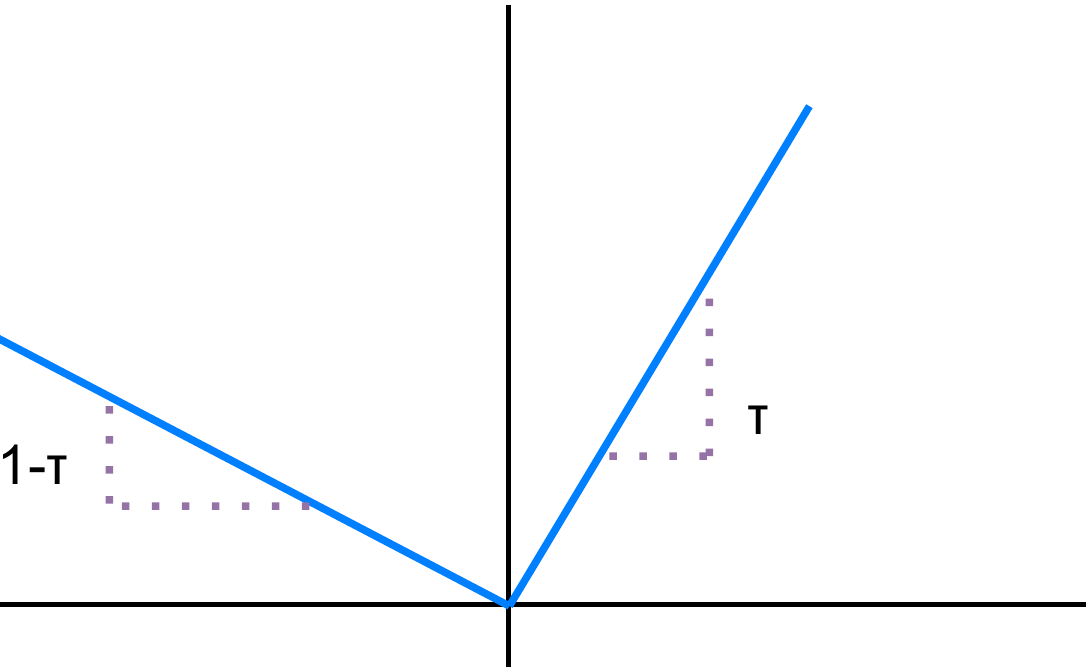}
    \caption{The pinball loss objective used for quantile regression. The slope of the lines is determined by the desired quantile level $\tau$.}
    \label{fig:pinball}
\end{figure}

Quantile regression is a statistical method used to model input-dependent quantiles within a regression framework \citep{koenker1978regression}. As opposed to regular (linear) regression that models the mean of a target variable $Y$ conditioned on the input $X$, quantile regression models a \textit{quantile} of the distribution of $Y$ (e.g. the median, the 95\%, or the 5\% percentile scores). By definition, quantile regression does not require any parametric assumptions on the distribution of $Y$ and is less sensitive to outliers. Quantiles provide an attractive representation for uncertainty: they allow for easy construction of prediction intervals, at chosen confidence levels. 
Learning the quantile for a particular quantile level involves
optimizing the \textbf{pinball loss}, a tilted transformation of the absolute value function (see Figure \ref{fig:pinball}). Given a
target $y$, a prediction $\hat{y}$, and quantile level $\tau \in (0,1)$, the pinball loss $\mathcal{L}_\tau$ is defined as: 
\begin{equation}
\label{eq:pinball}
    \mathcal{L}_\tau(\hat{y}; y) = (\hat{y}-y)(\mathds{1}\{y \leq \hat{y}\}-\tau).
\end{equation}

We can select $\tau$ to correspond to the error rate $\alpha$ that we want to achieve. Note that for  $\tau=0.5$  the  the loss function reduces to (half) the mean absolute error loss $\mathcal{L}_\mathrm{MAE}(\hat{y}; y) = \frac{1}{2} {|\hat{y}-y|}$.

We use $\tau=\alpha$ to train our models to predict the $\hat{Q}_{1-\tau/2}$ and $\hat{Q}_{\tau/2}$ quantiles, as well as the $\hat{Q}_{0.5}$ quantile, which corresponds to the median (see below), but there are extensions that either optimize multiple quantiles that cover the full predictive distribution \cite{tagasovska2019single} or explore asymmetric loss extensions to account for overestimating or underestimating the confidence intervals \cite{beck2016exploring}.


Unlike the parametric methods covered in \S\ref{sec:parametric}, the quantile regression method can be used to return \textbf{asymmetric} confidence intervals. 
This is done by fitting 0.5, $1-\frac{\tau}{2}$, and $\frac{\tau}{2}$ quantile predictors to the data, and setting $\hat{y}(x) := \hat{Q}_{0.5}(x)$, $\hat{\delta}_+(x) := \hat{Q}_{1-\frac{\tau}{2}}(x) - \hat{y}(x)$, and $\hat{\delta}_-(x) := \hat{y}(x) - \hat{Q}_{\frac{\tau}{2}}(x)$. 

For completeness, we also consider a \textbf{symmetric} variant of quantile regression where we do not estimate the median $\hat{Q}_{0.5}(x)$, but rather set $\hat{y}(x) = \frac{1}{2}\big(\hat{Q}_{1-\frac{\tau}{2}}(x)+\hat{Q}_{\frac{\tau}{2}}(x)\big)$. 
We report coverage for both the non-symmetric (\textbf{QNT-NS}) and the symmetric case (\textbf{QNT-S}) later in Table~\ref{tab:coverage}.


\subsection{Comparison with calibration}\label{sec:calib}
We also compare the coverage obtained by uncertainty methods and conformal prediction to  a calibration approach that aims to minimize the \textbf{expected calibration error} (ECE; \cite{naeini2015obtaining,pmlr-v80-kuleshov18a}). ECE has been proposed as a measure of how well aligned the model confidence is with the model accuracy, based on the simple desideratum that a model with e.g. 80\% confidence over a set of examples should achieve an accuracy of 80\% over the same examples to be well-calibrated. It is defined as 
\begin{align}
\label{eq:ece}
    \mathrm{ECE} &= \frac{1}{M}\sum_{b=1}^{M} |\mathrm{acc}(\gamma_{b}) - \gamma_{b}|,
\end{align}
where each $b$ is a bin representing a confidence level $\gamma_b$, and $\mathrm{acc}(\gamma_{b})$ is the fraction of times the ground truth $y$ falls inside the confidence interval associated to that bin. 
Several variants of uncertainty calibration have been proposed to correct unreliable uncertainty estimates that do not correlate with model accuracy \cite{pmlr-v80-kuleshov18a,amini2020deep,levi2022evaluating}. We follow \citet{glushkova2021uncertainty} who find that computing a simple affine transformation of the original uncertainty distribution that minimises the ECE is effective to quantify uncertainty in MT evaluation.

\subsection{Results}
\label{sec:res_unc}
We first compare the uncertainty methods described in \S\ref{sec:unc_meth} with respect to coverage percentage as shown in Table \ref{tab:coverage}. We select a desired coverage level of $90\%$, i.e., we set $\alpha=0.1$. We also align the uncertainty estimates with respect to the same $\alpha$ value: for the parametric uncertainty heuristics, we select the $\delta(x)$ that corresponds to a $1-\alpha$ coverage of the distribution, by using the probit function as described in \S\ref{sec:parametric}; and for the non-parametric approach, we train the quantile regressors by setting $\tau = \alpha/2$, as described in \S\ref{sec:quantile}. 

\begin{table}[t]
    \centering
    \begin{tabular}{ccc cc}
    \toprule
         Method & Orig. & Calib. & Conform.   & $\hat{q}$ 
         \\
    \midrule     
         QNT-NS &  77.83 &  --   & 90.21 & 1.44  \\
         QNT-S  &  78.66 & 69.03 & 90.54 & 1.41  \\
         MCD & 23.82 & 66.60 & 90.01 & 8.01 \\
         DE  & 29.10 & 66.23 & 91.31 & 7.06 \\
         HTS & 82.18 & 68.29 & 89.89 & 1.31 \\
         DUP & 86.23 & 66.13 & 89.88 & 1.14 \\
    \bottomrule
    \end{tabular}
    \caption{Coverage percentage for $\alpha = 0.1$ over different uncertainty methods. Values reported correspond to the mean over 10 runs. The second, third, and fourth columns refer respectively to the coverage obtained by original methods without calibration, after the ECE calibration described in \S\ref{sec:calib}, and with the conformal prediction procedure described in \S\ref{sec:conformal}.}
    \label{tab:coverage}
\end{table}

Table~\ref{tab:coverage} shows that coverage varies significantly across methods, with the sampling-based methods such as MC dropout and deep ensembles achieving coverage much below the desired $1-\alpha$ level. In contrast, direct uncertainty prediction achieves comparatively high coverage even before the application of conformal prediction. This could be related to the fact that by definition, the DUP method tries to predict uncertainty modeled as $\epsilon = |\hat{y}-y|$.

Calibration helps improve coverage in the cases of MC dropout and deep ensembles---albeit still without reaching close to 0.9. Instead, it seems that minimizing the ECE is not well aligned to optimizing coverage as for all cases calibration leads to less than $70 \%$ coverage.
In contrast, we can see that conformal prediction approximates the desired coverage level best for all methods, regardless of the initial coverage they obtain, in line with the guarantees provided by Theorem~\ref{theo:conf}.

Besides measuring the coverage achieved by the several methods, it is also important to examine the \textit{width} of the predicted intervals---if intervals are too wide, they will not be very informative nor  useful in practice (see also Figure \ref{fig:cover_ideal}). To that end, we also compute the \textbf{sharpness} \citep{pmlr-v162-kuleshov22a} as the average width of the predicted confidence intervals after the conformal prediction application~\footnote{In related work \cite{glushkova2021uncertainty} sharpness is computed with respect to $\sigma^2$, but this cannot be applied to non-parametric uncertainty cases, so we use the confidence interval length to be able to compare conformal prediction for all uncertainty quantification methods.}, 
\begin{equation}
    \mathrm{sha} = \frac{1}{|\mathcal{S}^{\mathrm{test}}|} \sum_{x \in \mathcal{S}^{\mathrm{test}}}  |\mathcal{C}_{\hat{q}}(x)|.
    \label{eq:sha}
\end{equation}
We show the results in Table \ref{tab:sha}. We observe that, for the majority of methods, the sharpness values are similar, except for deep ensembles (DE) where we need to rely on wider intervals to achieve the same level of coverage.


\begin{table}[t]
    \centering
    \begin{tabular}{cc}
   \toprule
    Method & Sharpness\\
    \midrule
    QNT-NS/S &    2.409\\
    MCD                  &    2.629\\
    DE                   &    3.074\\
    HTS                  &    2.304\\
    DUP                  &    2.486\\
    \bottomrule
    \end{tabular}
    \caption{Sharpness of confidence intervals for each uncertainty quantification method after conformal prediction.}
    \label{tab:sha}
\end{table}

\section{Conditional Coverage}
\label{sec:conditional}

The coverage guarantees stated in Theorem~\ref{theo:conf} refer to \textit{marginal} coverage---the probabilities are not conditioned on the input points, they are averaged (marginalized) over the full test set. 
In several practical situations it is  desirable to assess the \textbf{conditonal} coverage $\mathds{P} [Y_{\mathrm{test}} \in \mathcal{C}(X_{\mathrm{test}}) \mid X_{\mathrm{test}} \in \mathcal{G}]$ where $\mathcal{G} \subseteq \mathcal{X}$ denotes a region of the input space, e.g., inputs containing some specific attributes or pertaining to some group of the population. 



\begin{figure}[h!]
    \begin{minipage}{\columnwidth}
    \includegraphics[trim={2.5cm 3.5cm 0. 2cm},clip,width=1.1\columnwidth]{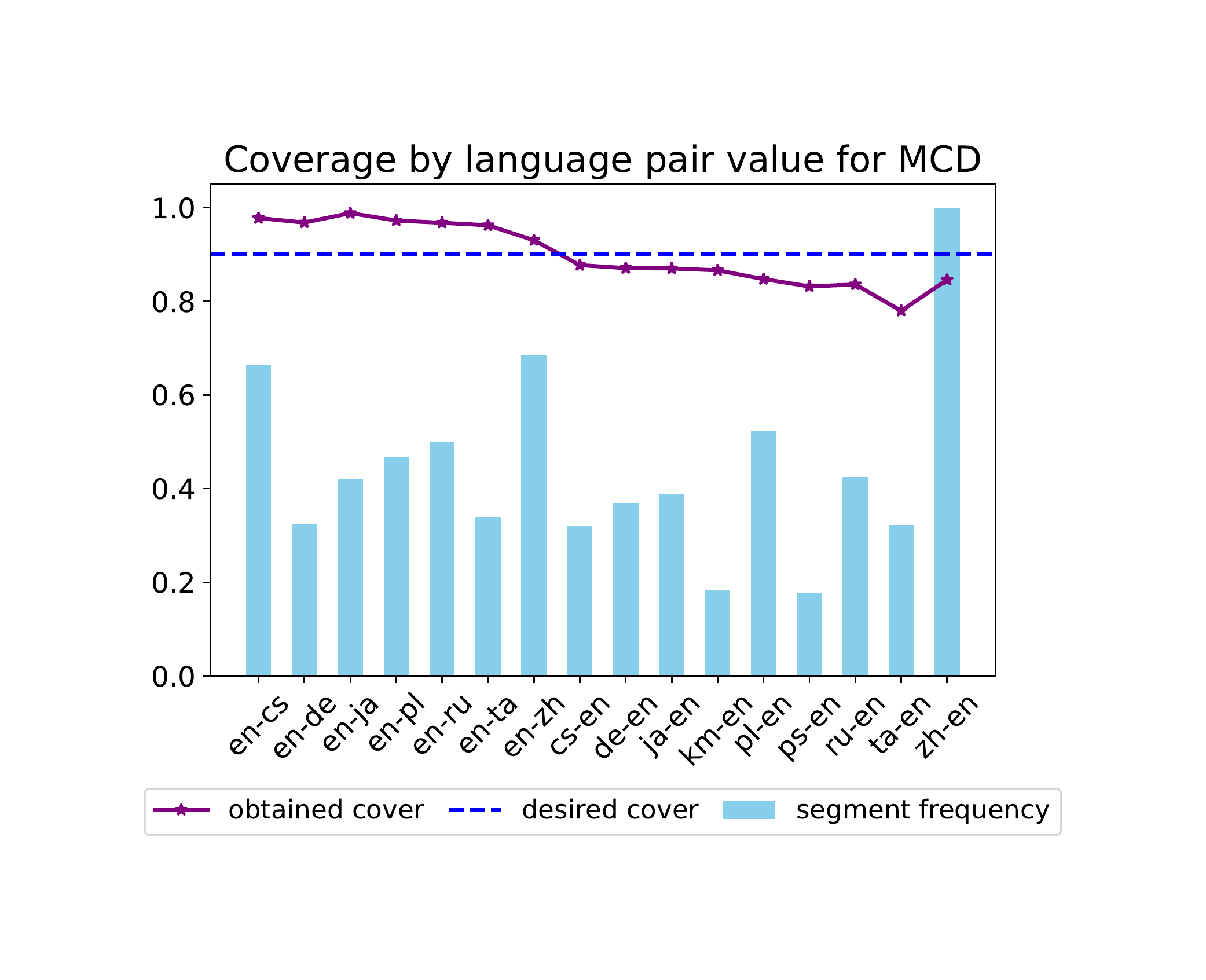}
    \end{minipage}
    \begin{minipage}{\columnwidth}
    \includegraphics[trim={2.5cm 3.5cm 0. 2cm},clip,width=1.1\columnwidth]{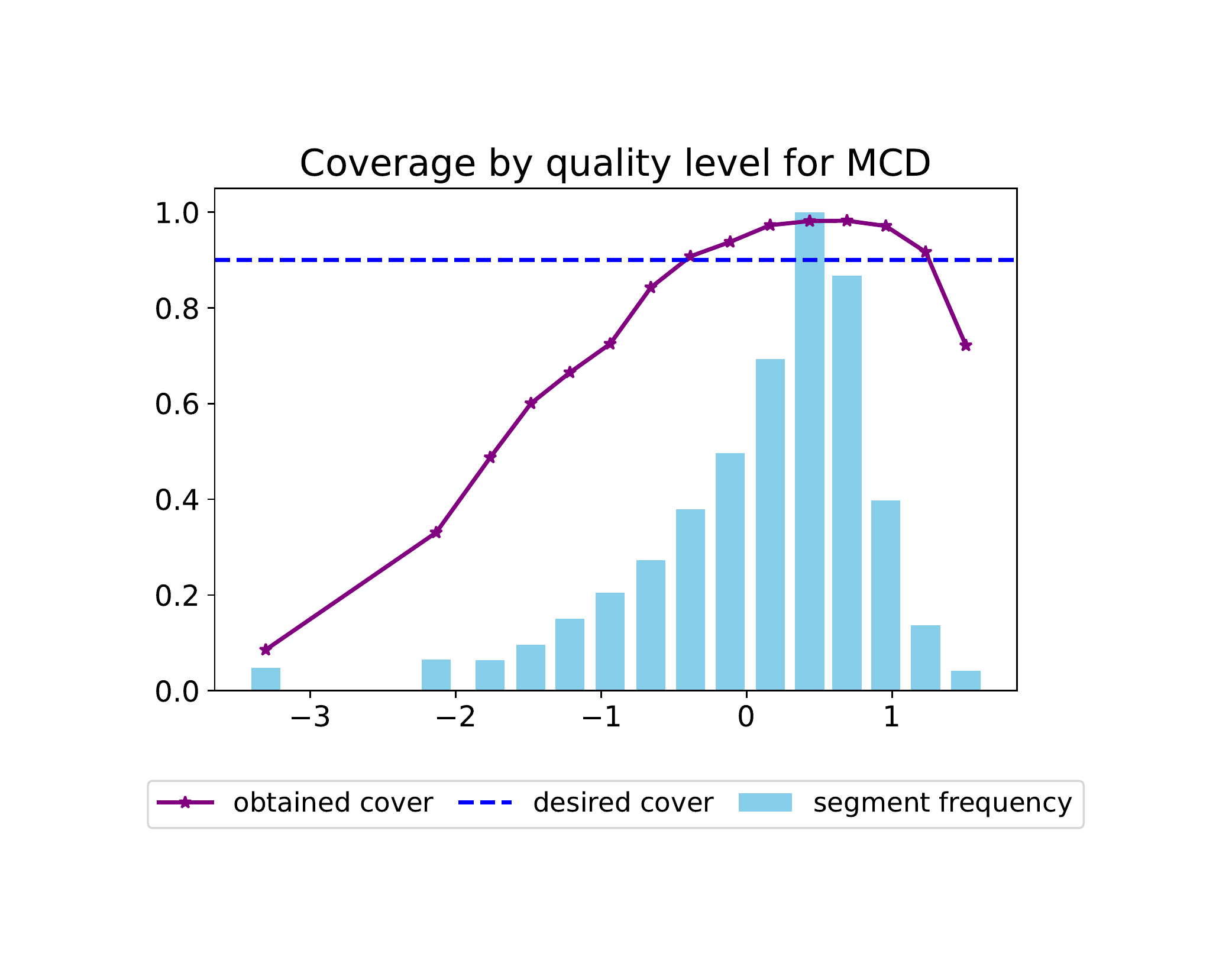}
    \end{minipage}
    \begin{minipage}{\columnwidth}
    \includegraphics[trim={2.5cm 1.5cm 0. 2cm},clip,width=1.1\columnwidth]{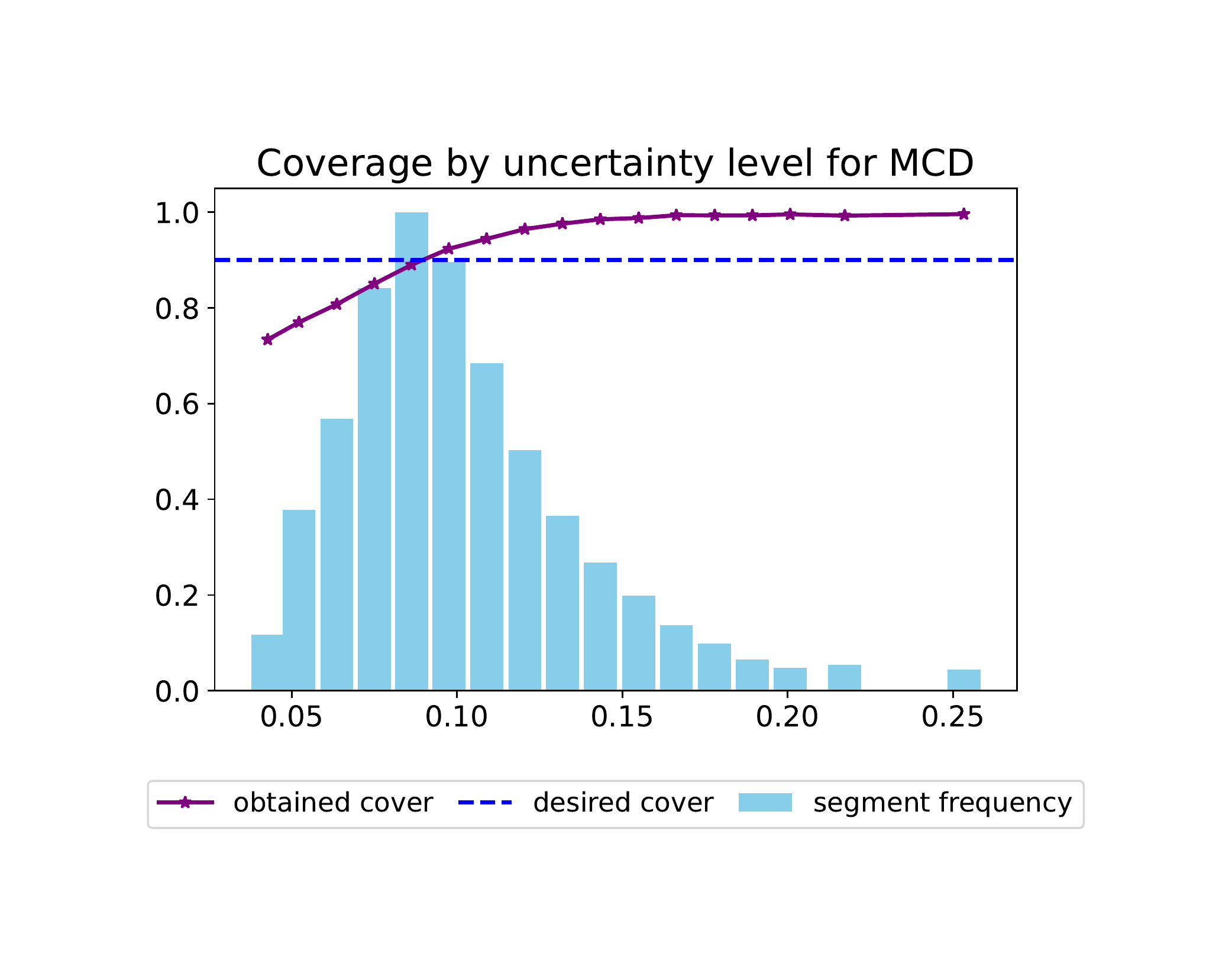}
    \end{minipage}
    \caption{Conditional coverage imbalance per language pair (top), quality score level (middle) and uncertainty score (bottom) for conformal prediction with MCD-based non-conformity scores \footnotemark. 
    }
    \label{fig:cover_imbalance}
\end{figure}

In fact, evaluating the conditional coverage with respect to different data attributes may reveal biases of the uncertainty estimation methods towards specific data subgroups which are missed if we only consider marginal coverage. In the next experiments, we follow the feature stratified coverage described  in~\citet{angelopoulos2021gentle}; we use conformal prediction with MC dropout as our main paradigm where MC dropout is 
used as the underlying uncertainty quantification method. We demonstrate three examples of imbalanced coverage in Figure \ref{fig:cover_imbalance} with respect to three different attributes: language pairs, translation quality level, and estimated uncertainty scores. 

We can see that, in all cases, coverage varies significantly across groups, revealing biases towards specific attribute values. For example, the plots show that into-English translations are under-covered (coverage $\leq 0.9$), i.e., we consistently underestimate the uncertainty over the predicted quality for these language pairs. More importantly, we can see that examples with low translation quality (measured by human scores) are significantly under-covered, as coverage for quality scores where $y \leq -1.5$ drops below 50\%. Observing the frequencies of these segments, we can see that the drop in segment frequency seems to correlate with a significant drop in coverage. For MCD-based uncertainty scores on the other hand, the drop in coverage seems to be related to the low uncertainty scores, indicating that due to the skewed distribution of uncertainty scores, the calculation of the $\hat{q}$ quantile is not well tuned to lower uncertainty values (i.e., higher non-conformity scores). Similar patterns for these three dimensions are also observed for the other uncertainty quantification methods revealing the biases of these methods. 

Ensuring we do not overestimate confidence for these examples (specially for low quality segments) is crucial for MT evaluation, in particular for applications where MT is used on the fly and one needs to know if a translation can be shared or would require further editing with human intervention. Hence, in the rest of this section, we elaborate approaches to assess and mitigate coverage imbalance in the aforementioned examples, in order to obtain more \textbf{equalized coverage} \cite{romano2020malice}. We consider  conditional coverage both for discrete (language pairs) and continuous (quality scores) attributes.

\footnotetext{Note that to facilitate plotting, the segment frequencies are re-scaled with respect to the maximum bin frequency (such that the bin with the maximum segment frequency  equals 1).  }

\subsection{Conditioning on discrete attributes: language-pairs}
\label{sec:lps}

To deal with imbalanced coverage for discrete data attributes, such as the language pair, we use an equalized conformal prediction approach, i.e., we compute the conditional coverage for each attribute value and, upon observing imbalances, \textbf{we compute conditional quantiles instead of a single one} on the calibration set. 

Let $\{1, \ldots, K\}$ index the several attributes (e.g., language pairs). 
We partition the calibration set according to these attributes,  $\mathcal{S}^{\textrm{cal}} = \bigcup_{k=1}^K \mathcal{S}_k^{\textrm{cal}}$, where $\mathcal{S}_k^{\textrm{cal}}$ denotes the partition corresponding to the $k$\textsuperscript{th} attribute and $\mathcal{S}_k^{\textrm{cal}} \cap \mathcal{S}_{k'}^{\textrm{cal}} = \varnothing$ for every $k \ne k'$. 
Then, we follow the procedure described in \S\ref{sec:conformal} to fit attribute-specific quantiles $\hat{q}_k$ to each calibration set $\mathcal{S}_k^{\textrm{cal}}$.




We demonstrate the application of this process on language pairs for all uncertainty quantification methods examined in the previous section. Table \ref{tab:lpcover} shows the language-based conditional coverage, using a heatmap coloring to highlight the language pairs that fall below the guaranteed marginal coverage of $1-\alpha=0.9$. We can see that for all language pairs we achieve coverage >75\% but some are below the 90\% target. For all methods except for DUP, the coverage is high for out-of-English translations and drops for the majority of into-English cases. Instead, DUP seems to be the method obtaining the most balanced performance across languages, with smaller deviations that do not seem to favor a specific translation direction. 
Applying the equalizing approach described above, we successfully rectify the imbalance for all uncertainty quantification methods, as shown in the heatmap of Table \ref{tab:lpcover2}.

\begin{table}[htb!]
\begin{tabular}{lrcccc}
\toprule
      & \multicolumn{1}{l}{QNT}  & MCD                           & DE                            & HTS                           & DUP                           \\ \midrule
En-Cs & \cellcolor[HTML]{FFFFE4}0.982 & \cellcolor[HTML]{FFFFE4}0.959 & \cellcolor[HTML]{FFFFE4}0.939 & \cellcolor[HTML]{FAD9C0}0.875 & \cellcolor[HTML]{FFFFE4}0.931 \\
En-De & \cellcolor[HTML]{FFFFE4}0.973 & \cellcolor[HTML]{FFFFE4}0.971 & \cellcolor[HTML]{FFFFE4}0.925 & \cellcolor[HTML]{F8C7B0}0.863 & \cellcolor[HTML]{FFFFE4}0.927 \\
En-Ja & \cellcolor[HTML]{FFFFE4}0.990 & \cellcolor[HTML]{FFFFE4}0.978 & \cellcolor[HTML]{FFFFE4}0.987 & \cellcolor[HTML]{FCE9D0}0.886 & \cellcolor[HTML]{FFFFE4}0.972 \\
En-Pl & \cellcolor[HTML]{FFFFE4}0.977 & \cellcolor[HTML]{FFFFE4}0.948 & \cellcolor[HTML]{FFFFE4}0.914 & \cellcolor[HTML]{FBE3CA}0.882 & \cellcolor[HTML]{FFFFE4}0.914 \\
En-Ru & \cellcolor[HTML]{FFFFE4}0.974 & \cellcolor[HTML]{FFFFE4}0.958 & \cellcolor[HTML]{FFFFE4}0.936 & \cellcolor[HTML]{F8C6AF}0.862 & \cellcolor[HTML]{FFFFE4}0.926 \\
En-Ta & \cellcolor[HTML]{FFFFE4}0.970 & \cellcolor[HTML]{FFFFE4}0.952 & \cellcolor[HTML]{FFFFE4}0.949 & \cellcolor[HTML]{FDF3D9}0.892 & \cellcolor[HTML]{F7BFA9}0.858 \\
En-Zh & \cellcolor[HTML]{FFFFE4}0.934 & \cellcolor[HTML]{FFFFE4}0.983 & \cellcolor[HTML]{FFFFE4}0.991 & \cellcolor[HTML]{FFFFE4}0.919 & \cellcolor[HTML]{FFFFE4}0.945 \\
Cs-En & \cellcolor[HTML]{FDEFD5}0.890 & \cellcolor[HTML]{FAD3BB}0.871 & \cellcolor[HTML]{FCE7CE}0.884 & \cellcolor[HTML]{FEFCE1}0.898 & \cellcolor[HTML]{FAD9C1}0.875 \\
De-En & \cellcolor[HTML]{FBE0C7}0.880 & \cellcolor[HTML]{FDEDD3}0.888 & \cellcolor[HTML]{F9CDB5}0.867 & \cellcolor[HTML]{FEF9DE}0.896 & \cellcolor[HTML]{FFFFE4}0.902 \\
Ja-En & \cellcolor[HTML]{FCE5CC}0.883 & \cellcolor[HTML]{F7BDA6}0.856 & \cellcolor[HTML]{FFFFE4}0.921 & \cellcolor[HTML]{FFFFE4}0.910 & \cellcolor[HTML]{FCEBD1}0.887 \\
Kn-En & \cellcolor[HTML]{FBE3CA}0.881 & \cellcolor[HTML]{FAD8C0}0.875 & \cellcolor[HTML]{FFFFE4}0.948 & \cellcolor[HTML]{FFFFE4}0.943 & \cellcolor[HTML]{F4A48F}0.840 \\
Pl-En & \cellcolor[HTML]{F8C5AE}0.862 & \cellcolor[HTML]{F39A86}0.833 & \cellcolor[HTML]{F28E7A}0.825 & \cellcolor[HTML]{FAD6BE}0.873 & \cellcolor[HTML]{F6B29D}0.849 \\
Ps-En & \cellcolor[HTML]{F6B59F}0.851 & \cellcolor[HTML]{F7B9A3}0.854 & \cellcolor[HTML]{FFFFE4}0.932 & \cellcolor[HTML]{FFFFE4}0.922 & \cellcolor[HTML]{EB5344}0.786 \\
Ru-En & \cellcolor[HTML]{F6B59F}0.851 & \cellcolor[HTML]{F2937F}0.828 & \cellcolor[HTML]{F39683}0.831 & \cellcolor[HTML]{FBDEC6}0.879 & \cellcolor[HTML]{FDEDD3}0.888 \\
Ta-En & \cellcolor[HTML]{ED5F4F}0.793 & \cellcolor[HTML]{EF7765}0.809 & \cellcolor[HTML]{FBDDC4}0.878 & \cellcolor[HTML]{FEFCE1}0.898 & \cellcolor[HTML]{FCE4CB}0.883 \\
Zh-En & \cellcolor[HTML]{F8C4AD}0.861 & \cellcolor[HTML]{F39A86}0.833 & \cellcolor[HTML]{F9CEB6}0.868 & \cellcolor[HTML]{FCE9CF}0.886 & \cellcolor[HTML]{F2917D}0.827\\ \bottomrule
\end{tabular}
\caption{\label{tab:lpcover}Conditional coverage over different language pairs of WMT 2020 DA data. Red coloured entries signify coverage < 0.9. }
\end{table}
\begin{table}[htb!]
\begin{tabular}{@{}lrcccc@{}}
\toprule
      & \multicolumn{1}{l}{QNT}       & MCD                           & DE                            & HTS                           & DUP                           \\ \midrule
En-Cs & \cellcolor[HTML]{FDF4D9}0.893 & \cellcolor[HTML]{FFFFE4}0.917 & \cellcolor[HTML]{FCECD3}0.888 & \cellcolor[HTML]{FDF3D9}0.892 & \cellcolor[HTML]{FFFFE4}0.902 \\
En-De & \cellcolor[HTML]{FFFFE4}0.902 & \cellcolor[HTML]{FFFFE4}0.902 & \cellcolor[HTML]{FFFFE4}0.902 & \cellcolor[HTML]{FEF8DD}0.896 & \cellcolor[HTML]{FDF3D9}0.893 \\
En-Ja & \cellcolor[HTML]{FFFFE4}0.909 & \cellcolor[HTML]{FDF1D7}0.891 & \cellcolor[HTML]{FFFFE4}0.900 & \cellcolor[HTML]{FDF2D7}0.891 & \cellcolor[HTML]{FFFFE4}0.904 \\
En-Pl & \cellcolor[HTML]{FCE4CB}0.882 & \cellcolor[HTML]{FFFFE4}0.905 & \cellcolor[HTML]{FEF7DD}0.895 & \cellcolor[HTML]{FFFFE4}0.900 & \cellcolor[HTML]{FEFBE1}0.898 \\
En-Ru & \cellcolor[HTML]{FEFEE3}0.900 & \cellcolor[HTML]{FEFBE1}0.898 & \cellcolor[HTML]{FFFFE4}0.908 & \cellcolor[HTML]{FFFFE4}0.906 & \cellcolor[HTML]{FFFFE4}0.903 \\
En-Ta & \cellcolor[HTML]{FFFFE4}0.903 & \cellcolor[HTML]{FEF7DD}0.895 & \cellcolor[HTML]{FCE4CB}0.883 & \cellcolor[HTML]{FCEAD0}0.886 & \cellcolor[HTML]{FFFFE4}0.903 \\
En-Zh & \cellcolor[HTML]{FBE0C7}0.880 & \cellcolor[HTML]{FDF0D6}0.890 & \cellcolor[HTML]{FCE7CE}0.884 & \cellcolor[HTML]{FEF8DD}0.896 & \cellcolor[HTML]{FEF9DE}0.896 \\
Cs-En & \cellcolor[HTML]{FDEFD5}0.890 & \cellcolor[HTML]{FFFFE4}0.917 & \cellcolor[HTML]{FFFFE4}0.909 & \cellcolor[HTML]{FFFFE4}0.904 & \cellcolor[HTML]{FDF5DB}0.894 \\
De-En & \cellcolor[HTML]{FEFBE0}0.897 & \cellcolor[HTML]{FFFFE4}0.901 & \cellcolor[HTML]{FFFFE4}0.901 & \cellcolor[HTML]{FEFADF}0.897 & \cellcolor[HTML]{FFFFE4}0.903 \\
Ja-En & \cellcolor[HTML]{FEFEE3}0.900 & \cellcolor[HTML]{FFFFE4}0.912 & \cellcolor[HTML]{FEFDE2}0.899 & \cellcolor[HTML]{FDF5DB}0.894 & \cellcolor[HTML]{FFFFE4}0.902 \\
Kn-En & \cellcolor[HTML]{FEF9DE}0.896 & \cellcolor[HTML]{FFFFE4}0.903 & \cellcolor[HTML]{FFFFE4}0.902 & \cellcolor[HTML]{FFFFE4}0.904 & \cellcolor[HTML]{FDF5DB}0.894 \\
Pl-En & \cellcolor[HTML]{FFFFE4}0.900 & \cellcolor[HTML]{FFFFE4}0.905 & \cellcolor[HTML]{FDF4DA}0.893 & \cellcolor[HTML]{FDF5DB}0.894 & \cellcolor[HTML]{FBDCC3}0.877 \\
Ps-En & \cellcolor[HTML]{FFFFE4}0.905 & \cellcolor[HTML]{FEFEE3}0.899 & \cellcolor[HTML]{FFFFE4}0.900 & \cellcolor[HTML]{FCE6CD}0.884 & \cellcolor[HTML]{FFFFE4}0.907 \\
Ru-En & \cellcolor[HTML]{FFFFE4}0.910 & \cellcolor[HTML]{FEF8DD}0.896 & \cellcolor[HTML]{FFFFE4}0.907 & \cellcolor[HTML]{FFFFE4}0.900 & \cellcolor[HTML]{FEFEE3}0.900 \\
Ta-En & \cellcolor[HTML]{FCE6CD}0.884 & \cellcolor[HTML]{FFFFE4}0.901 & \cellcolor[HTML]{FCEAD0}0.886 & \cellcolor[HTML]{FFFFE4}0.901 & \cellcolor[HTML]{FFFFE4}0.908 \\
Zh-En & \cellcolor[HTML]{FEFEE3}0.900 & \cellcolor[HTML]{FFFFE4}0.910 & \cellcolor[HTML]{FFFFE4}0.908 & \cellcolor[HTML]{FEFEE3}0.900 & \cellcolor[HTML]{FFFFE4}0.905 \\ \bottomrule
\end{tabular}
\caption{Conditional coverage over different language pairs of WMT 2020 DA data, after balanced conformal prediction. Red coloured entries signify coverage < 0.9. }
\label{tab:lpcover2}
\end{table}

\subsection{Conditioning on continuous attributes: translation quality and uncertainty scores}\label{sec:mondrian}

With some additional constraints on the equalized conformal prediction process described in \S\ref{sec:lps} we can generalize this approach to account for attributes with continuous values, such as the quality scores (ground truth quality $y$) in the case of MT evaluation or the uncertainty scores obtained by different uncertainty quantification methods. To that end, we adapt the  \textbf{Mondrian conformal prediction} methodology \cite{vovk2005algorithmic}. 
Mondrian conformal predictors have been used initially for classification and later for regression, where they have been used to partition the data with respect to the residuals $|y-\hat{y}(x)|$ \cite{johansson2014regression}. \citet{bostrom2020mondrian} proposed a Mondrian conformal predictor that partitions along the expected ``difficulty'' of the data as estimated by the non-conformity score $s(x,y)$ or the uncertainty score $\delta(x)$. 

In all the above cases, the calibration instances are sorted according to continuous variable of interest and then partitioned into calibration bins. While the bins do not need to be of equal size, they need to satisfy a minimum length condition that depends on the chosen $\alpha$ threshold for the error rate \cite{johansson2014regression}. 
Upon obtaining a partition into calibration bins, and similarly to what was described in \S\ref{sec:lps} for discrete attributes, 
we compute bin-specific quantiles $\hat{q}_b$, where $b \in \{1, \ldots, B\}$ indexes a bin. 

We apply the aforementioned approach to the MT evaluation for the translation quality scores as well as the predicted uncertainty scores. For both cases, we set a threshold of at least $200$ segments per bin. For the \textbf{translation quality scores}, we initially compute quality score bins and quantiles over the calibration set using the ground truth scores $y$ as described previously. Subsequently, to apply the conformal prediction on a test instance $x_\mathrm{test}$, we use the predicted quality score $\hat{y}(x_\mathrm{test})$ and choose the best bin $\hat{b}$ to use by computing the smallest difference between $\hat{y}(x_\mathrm{test})$ and the mean quality score $\bar{y}_b$ of the calibration bins. 
We then predict the confidence interval by using the corresponding quantile $\hat{q}_{\hat{b}}$.  
For the \textbf{uncertainty scores}, the application is easier, since we have access to the uncertainty prediction for both the calibration and test instances. Thus upon splitting the calibration set with respect to the uncertainty scores and computing the new quantiles $\hat{q}_b$  per bin, we can directly identify the bin that $\delta(x_\mathrm{test})$ falls into. 

\begin{figure}[ht!]
    \centering
    \includegraphics[trim={2.5cm 1.5cm 0. 2.5cm},clip,width=1.1\columnwidth]{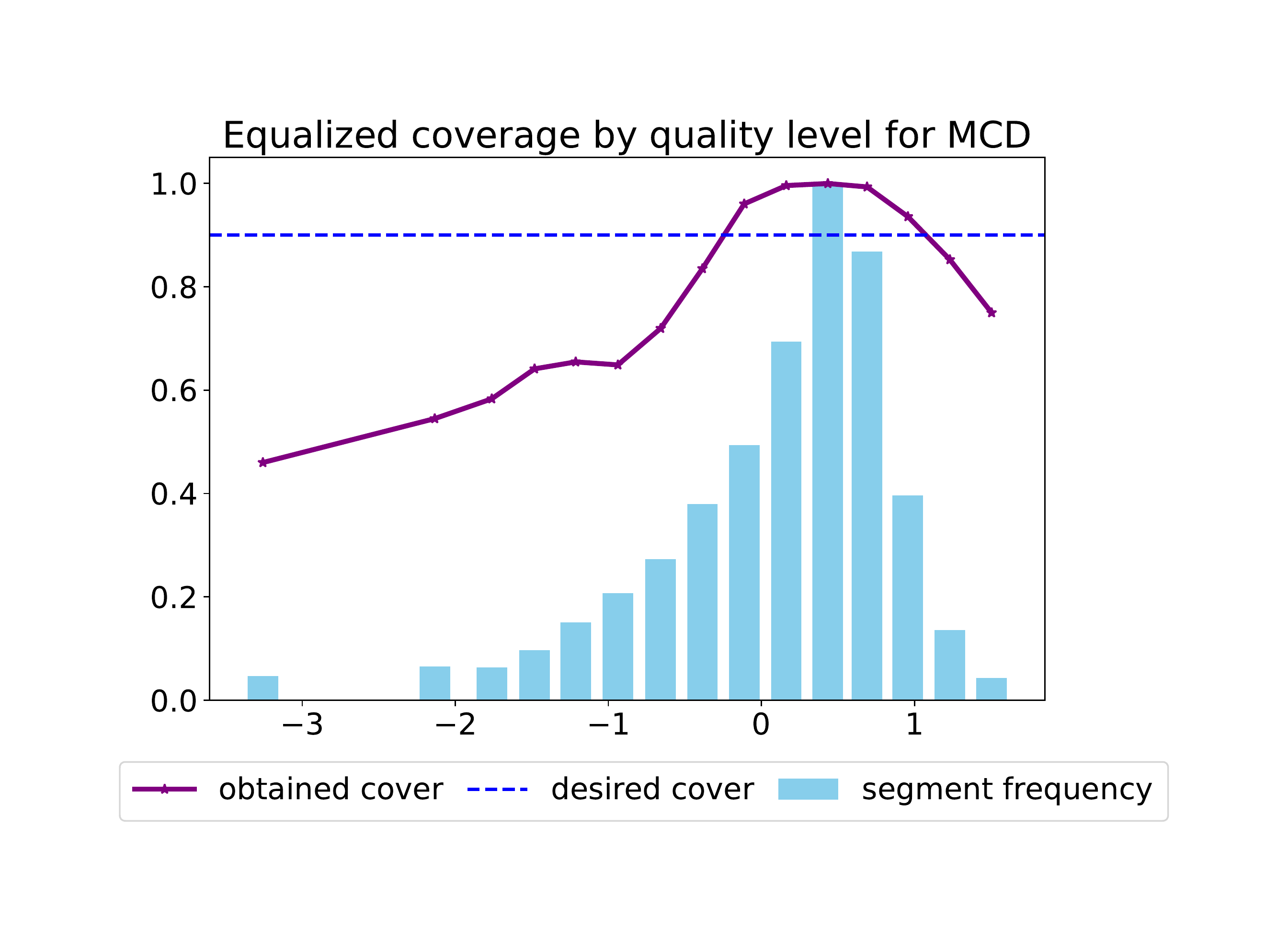}
    \caption{Equalized coverage over translation quality scores.}
    \label{fig:eq_cover_qual}
\end{figure}

\begin{figure}[ht!]
    \centering
    \includegraphics[trim={2.5cm 1.5cm 0. 2.5cm},clip,width=1.1\columnwidth]{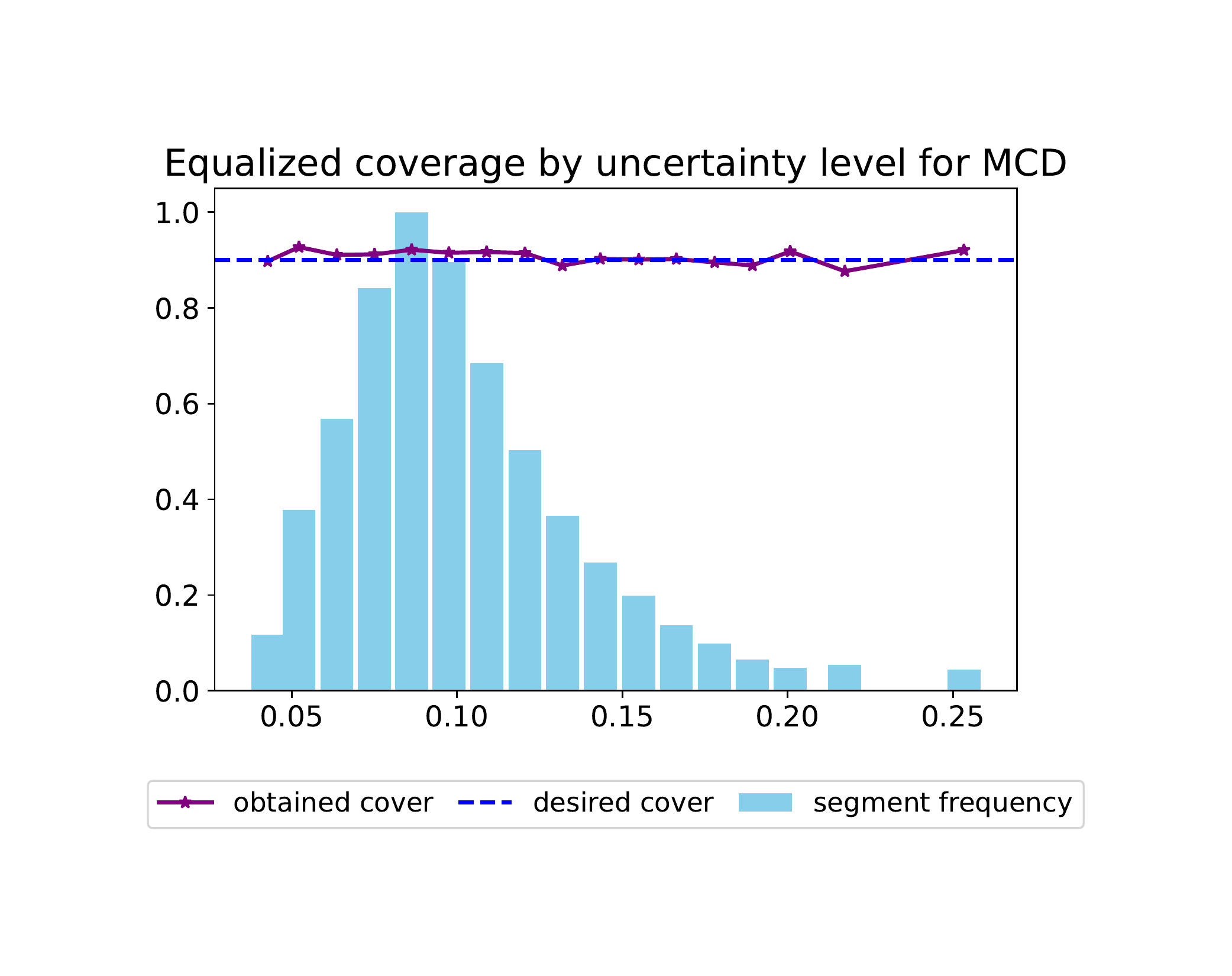}
    \caption{Equalized coverage over uncertainty scores.}
    \label{fig:eq_cover_unc}
\end{figure}

The equalized coverage over binned human scores and uncertainty values is shown in Figures \ref{fig:eq_cover_qual} and \ref{fig:eq_cover_unc}, respectively. We can see that in both cases the coverage approaches the threshold better (cf.~the original coverage in Figure~\ref{fig:cover_imbalance}), but not equalized performance is not equally achievable for both cases. For the case of translation uncertainty we manage to achieve balanced coverage across bins, that is very closed to the desired one. Instead, for the case of translation quality, we can see that the coverage for the very low quality translations is increased compared to the original coverage but is still far away from the desired threshold (similar behaviour can be observed for the very high quality translations). We hypothesize that this behaviour relates to our approximation of quality bin on the test data using the model predictions instead of the ground truth values. 


\section{Related Work}

\subsection{Conformal prediction}
We build on literature on conformal prediction that has been established by \citet{vovk2005algorithmic}, proposed as a finite-sample, distribution-free method for obtaining confidence intervals with guarantees on a new sample. Subsequent works focus on improving the predictive efficiency of the conformal sets or relaxing some of the constraints \cite{angelopoulos2021gentle,jin2022selection,tibshirani2019conformal}. Most relevant to our paper are works that touch conformal prediction for regression tasks, either via the use of quantile regression \cite{romano2019conformalized} or using other scalar uncertainty estimates \cite{angelopoulos2021gentle,johansson2014regression,papadopoulos2011regression}. Other strands of work focus on conditional conformal prediction and methods to achieve balanced coverage across different attributes (also referred to as equalized coverage)  \cite{angelopoulos2021gentle,romano2020malice,bostrom2021mondrian,lu2022fair}. 

There are few works that use conformal prediction in NLP, so far focusing only on classification or generation. Specifically, there have been some attempts to apply conformal prediction to sentence classification tasks, such as sentiment and relation classification and entity detection \cite{fisch21fewshot,fisch2022conformal,maltoudoglou2020bert}. \citet{maltoudoglou2020bert} use a transformer-based architecture to classify sentiment and then use the output probabilities of classifier to compute non-conformity scores and obtain prediction sets via conformal prediction. \citet{fisch2022conformal}, on the other hand, focus on obtain tight prediction sets while maintaining the marginal coverage guarantees by casting conformal prediction as a meta-learning paradigm over exchangeable collections of auxiliary tasks, applied on both text and image classification. Recently, \citet{ravfogel2023conformal} and \citet{kumar2023conformal} considered natural language generation, with the former proposing the use of conformal prediction applied to top-$p$ nucleus sampling, and the latter proposing the use of conformal prediction to quantify the uncertainty of large language models for question answering. They specifically proposed the use of conformal prediction to calibrate the parameter $p$ as a function of the entropy of the next word distribution. Concurrent to this work, \citet{giovannotti2023evaluating} proposed the use of conformal prediction with $k$-nearest neighbor ($k$NN) non-conformity scores as a method to quantify uncertainty for MT quality estimation, and show that it can be used as a new standalone uncertainty quantification method for this task. They empirically demonstrate the impact of violating the i.i.d. assumption on the obtained performance and show that $k$NN conformal prediction outperforms a fixed-variance baseline with respect to ECE, AUROC and sharpness, but they do not consider the aspect of marginal or conditional coverage for the estimated confidence intervals. They also did not consider any of the uncertainty quantification methods discussed in this paper as non-conformity scores.

Our work complements the aforementioned efforts, as it focuses on a regression NLP task (MT evaluation) and investigates the impact of conformal prediction on the estimated confidence intervals. Contrary to previous approaches, however, we provide a detailed analysis of conformal prediction for an NLP regression task and  investigate a wide range of uncertainty methods that can be used to design non-conformity scores. Additionally, we elaborate different aspects of equalized coverage for MT evaluation, revealing biases with respect to different data attributes, and providing an effective method that corrects for these biases.

\subsection{Uncertainty quantification}
Several uncertainty methods have been previously proposed for regression tasks in NLP and the task of MT evaluation specifically. \citet{beck2016exploring} focused on the use of Gaussian processes to obtain uncertainty predictions for the task of quality estimation, with emphasis on cases of asymmetric risk. \citet{wang2022uncertainty} also explored Gaussian processes, but provided a comparison of multiple NLP regression tasks (semantic sentence similarity, MT evaluation, sentiment quantification) investigating end-to-end and pipeline approaches to apply Bayesian regression to large language models. Focusing on MT evaluation, \citet{glushkova2021uncertainty} proposed the use of MC dropout and deep ensembles as efficient approximations of Bayesian regression, inspired by work in computer vision \cite{kendall2017uncertainties}. \citet{zerva2022disentangling} proposed additional methods of uncertainty quantification for MT evaluation, focusing on methods that target aleatoric or epistemic uncertainties under specific assumptions. They specifically investigated heteroscedastic regression and KL-divergence for aleatoric uncertainty and direct uncertainty prediction for epistemic uncertainty, highlighting the performance benefits of these methods, when compared to MC dropout and deep ensembles, with respect to correlation of uncertainties to model error. However, none of the previous works in uncertainty for NLP regression considered the aspect of coverage. We compare several of the aforementioned uncertainty quantification methods with respect to coverage and focus on the impact of applying conformal prediction to each uncertainty method.


\section{Conclusions}
In this work, we apply conformal prediction to the important problem of MT evaluation. We show that most existing uncertainty quantification methods significantly underestimate uncertainty, achieving low coverage, and that the application of conformal prediction can help rectify this and guarantee coverage tuned to a user-specified threshold. We also use conformal prediction tools to assess the conditional coverage for three different attributes: language pairs, translation quality, and estimated uncertainty level. We highlight inconsistencies and imbalanced coverage in all three cases, and we show that equalized conformal prediction can correct the initially unfair confidence predictions to obtain more balanced coverage across attributes. 

Overall, our work aims to highlight the potential weaknesses of using uncertainty estimation methods without a principled calibration procedure. To this end, we propose a methodology that can guarantee more meaningful confidence intervals. In future work, we aim to further investigate the application of conformal prediction across different data dimensions as well as different regression tasks in NLP.  \looseness=-1
\bibliography{tacl2021}
\bibliographystyle{acl_natbib}

\onecolumn

\appendix

\end{document}